\documentclass{article}

    \usepackage[T1]{fontenc}
    \usepackage{microtype}
    \usepackage{graphicx}
    \usepackage{subfigure}
    \usepackage{booktabs} 
    
    \PassOptionsToPackage{hyphens}{url}\usepackage{hyperref}
    \usepackage[hyphenbreaks]{breakurl}

    \usepackage[accepted]{icml2024}
    
    \usepackage{amsmath}
    \usepackage{amssymb}
    \usepackage{mathtools}
    \usepackage{amsthm}
    
    \usepackage[capitalize,noabbrev]{cleveref}

    \usepackage[utf8]{inputenc}
    \usepackage[switch]{lineno}
    \usepackage{commath}
    \usepackage{siunitx}
    \usepackage{eurosym}
    \usepackage[shortlabels]{enumitem}
    \usepackage[footnotehyper]{nicematrix}
    \usepackage{xspace}
    \usepackage{subcaption}
    \usepackage[flushleft]{threeparttable}
    \usepackage{multicol}
    \usepackage{wrapfig}
    
    \theoremstyle{plain}

    \theoremstyle{definition}

    \theoremstyle{remark}

    \usepackage[textsize=tiny]{todonotes}

    \def\aia{AI Act\xspace}

    \icmltitlerunning{FL and AI Regulation in the European Union: Who is Responsible? -- An Interdisciplinary Analysis}
    
\begin{document}
    
    \twocolumn[
    \icmltitle{Federated Learning and AI Regulation in the European Union: \\ Who is Responsible? -- An Interdisciplinary Analysis}

    \icmlsetsymbol{equal}{*}
    
    \begin{icmlauthorlist}
    \icmlauthor{Herbert Woisetschl\"ager}{equal,tum}
    \icmlauthor{Simon Mertel}{equal,ubtc}
    \icmlauthor{Christoph Kr\"onke}{ubtl}
    \icmlauthor{Ruben Mayer}{ubtc}
    \icmlauthor{Hans-Arno Jacobsen}{uot}
    \end{icmlauthorlist}

    \icmlaffiliation{tum}{School of Computation, Information and Technology, Technical University of Munich, Germany}
    \icmlaffiliation{ubtl}{Department of Law \& Economics, University of Bayreuth, Germany}
    \icmlaffiliation{ubtc}{Department of Computer Science, University of Bayreuth, Germany}
    \icmlaffiliation{uot}{Department of Electrical and Computer Engineering, University of Toronto, Canada}

    \icmlcorrespondingauthor{Herbert Woisetschl\"ager}{herbert.woisetschlaeger@tum.de}
    \icmlcorrespondingauthor{Simon Mertel}{simon.mertel@uni-bayreuth.de}
    \icmlcorrespondingauthor{Christoph Kr\"onke}{christoph.kroenke@uni-bayreuth.de}
    \icmlcorrespondingauthor{Ruben Mayer}{ruben.mayer@uni-bayreuth.de}
    \icmlcorrespondingauthor{Hans-Arno Jacobsen}{jacobsen@eecg.toronto.edu}
    
    \icmlkeywords{Federated Learning, Legal Analysis, Regulatory Responsibility}
    
    \vskip 0.3in
    ]
    
    
    
    \printAffiliationsAndNotice{\icmlEqualContribution} 
    
    \begin{abstract}
    The European Union Artificial Intelligence Act mandates clear stakeholder responsibilities in developing and deploying machine learning applications to avoid substantial fines, prioritizing private and secure data processing with data remaining at its origin. Federated Learning (FL) enables the training of generative AI Models across data siloes, sharing only model parameters while improving data security.
    Since FL is a cooperative learning paradigm, clients and servers naturally share legal responsibility in the FL pipeline.
    Our work contributes to clarifying the roles of both parties, explains strategies for shifting responsibilities to the server operator, and points out open technical challenges that we must solve to improve FL's practical applicability under the EU AI Act.

    \end{abstract}
    
    \section{Introduction}
    \label{sec:intro}
With the introduction of the European Union Artificial Intelligence Act (\aia) \cite{eu_ai_act} and other international regulations being on the horizon, e.g., in the United States \cite{wh_executive_order} and Canada \cite{canada_act}, everyone concerned with the development and deployment of AI has to adapt to new game rules.
This entails data governance, robustness against adversarial scenarios, and energy considerations \cite{woisetschlaeger2024}. 
The \aia puts the \emph{service provider} into the spotlight, who has to assume responsibility for model development and deployment within the meaning of Article 3.
Especially regarding data governance, the \aia instantiates extensive rules for high-risk and general-purpose AI applications (GPAI, Article 52) that cater to data privacy and system security. 
The majority of generative AI applications fall under the GPAI definition in Article 3.

Federated Learning (FL) presents a privacy-enhancing and data-protecting machine learning technique \cite{mcmahan2017} that has recently received increased attention for enabling access to data silos for generative AI applications \cite{Woisetschlaeger2024a}.
In FL, a server operator provides an ML model sent to several clients and then trained on the clients' local data, which collaboratively train a global model via a central server, aggregating their local model updates.
Private and secure computing techniques like Differential Privacy or Trusted Execution Environments help improve data privacy and system security \cite{Bonawitz2017, Andrew2019}.
FL's data locality removes the key challenge of monitoring data lineage and simplifies accounting for user consent.
Specifically, we study the FL workflow in alignment with related work \cite{Li2020, Hard2018, mcmahan2017} to touch up on the following: 

\textbf{Data Acquisition}. 
The server operator can only employ a variety of client sampling strategies \cite{malinovsky2023federated, wang2022a, mcmahan2017} for an FL training round, without the ability to directly investigate client data or process integrity.

\textbf{Data Storage}. Similarly, the clients decide how, where, and when to store data. 
This has implications on data availability, which directly touches upon the \aia data governance requirements (Article 10)\footnote{In the following, the term Article refers to articles in the \aia if not specified otherwise}.

\textbf{Data Preprocessing}. While the server operator can provide instructions on how to preprocess data so that the data is compatible with the ML model, the clients have the freedom to run additional preprocessing steps. 
Since the server operator has no direct data access, verifying data integrity before training is challenging. 
For FL applications, there are numerous approaches to improve data integrity \cite{SnchezSnchez2024, RoyChowdhury2022}.

\textbf{Model Aggregation}.
While acting as the FL training orchestrator, the server operator handles the model integrity control mechanism when aggregating model updates.
Thus, FL appears to be a well-suited solution to open up data silos and provide access to additional data. 
This would significantly benefit the training or fine-tuning of generative models due to their sheer appetite for ever-increasing amounts of data \cite{Zhou2023}. 

One can think that the server operator is automatically also the \emph{service provider}.
Yet, FL is a cooperative ML training technique where a central entity typically provides the ML model, and clients can decide when to participate and what data to use for training. 
As such, we see that the server (model) and the clients (data) control parts of the FL lifecycle, rendering them both legally responsible for their respective parts.
Thus, this opens up the question: 
\vspace{-6pt}
\begin{quote}
    \textit{Who is the service provider at what point in the FL workflow, and how can each party assume adequate responsibility?}
\end{quote}
\vspace{-6pt}

Our paper studies technical and legal requirements that need to be established so that the FL server operator can assume responsibility as a service provider. This requires future technical work on auditability, verifiability, integrity, and privacy.
Further, we need to establish regulatory references for the terms and services of FL applications.





    \vspace{-6pt}
    \section{Technical Solutions Need to Focus on Transferring Responsibility to the Server Operator}
    \label{sec:technical_solutions}
    \emph{For practical FL applications, 
the server operator must assume the role of the service provider by employing appropriate
technical solutions.}

When establishing an FL system that could potentially entail thousands of clients at a time, managing responsibilities is likely to become a key challenge. Thus, we require solutions that provide for auditability, verifiability, integrity, and privacy. 

\textbf{Auditability \& verifiability}. 
There is a natural trade-off between privacy and data audits. 
The core paradigm of FL is to not share data beyond a client's area of control. Data is strictly inaccessible for everybody but the data owner, and even in-house restricted to authorized personnel. 
Thus, we face a challenge when aiming to audit all steps that happen on a client device or data server. 
For instance, a work by \citep{Liu2023} uses a Bayesian Nash equilibrium and a market mechanism to incentivize truthful client behavior, i.e., submission of useful model updates. 
While this approach significantly reduces the risk of adversarial attacks, it does not meet the requirements for auditing in the context of the \aia, which are well-defined. 
Quintessentially, any data that is being captured, processed, and used in a training process must be evaluated for potential bias or adversarial information.
To achieve this, numerous works combining FL with blockchain technology explore auditing the data processing steps and the training itself \cite{Nguyen2021, Ma2020}. 
What remains open is to develop solutions against data tampering.

\textbf{Integrity \& privacy}. 
Particularly, we have to rethink the obligations of the provider concerning data integrity and protection (Articles 8--10), such that responsibility is transferred to the FL server. 
To account for the asymmetry of access and control-by-design in FL systems, we must develop data integrity measures that capture the nature of client data at the time of collection, while preprocessing the data, and immediately before starting the training process.
Peer-based verification schemes of model updates are a promising direction to identify adversarial clients \cite{RoyChowdhury2022}. 
Extending such schemes from client models to client data without infringing privacy would be interesting.  At the same time, technical solutions must be in line with the requirements set out in the GDPR, which are not (necessarily) aligned with the concepts and rules of the \aia.


    \vspace{-6pt}
    \section{Regulatory Implementations Need to Foster Integrity and Verifiability}
    \label{sec:regulatory_setup}
\textit{We need FL server operators to assume full responsibility; clients are technically and legally obligated to comply.}

\textbf{Service Provider.} 
The GDPR \cite{gdpr2022} defines the term \emph{data controller}. Complementary, the \aia defines the \emph{service provider} of AI systems.
For data protection assessment when processing personal information at first, we need to clarify who is the data controller responsible and accountable for each distinct phase of the data processing and must demonstrate compliance with the requirements of the GDPR (Article 5). The \aia does not have a differentiated allocation of roles for separate processing phases and focuses on one central “provider” of a (compliant) AI system, defined in Article 3, with the obligations arising from Article 8.

While both the FL server and clients could be considered providers under the \aia, since the \aia (unlike the GDPR) focuses less on responsibilities for individual, definable data processing phases and more on secure system design as a whole, the provider concept has to be teleologically limited to the FL server. 
Thus, the server acts as the fully responsible service provider under the \aia (Article 8), especially concerning data governance (Article 10) and General-Purpose AI service (Article 52).

\textbf{General Terms and Conditions for AI Systems.} 
While in Article 4 of the GDPR, the controller is the person who, alone or jointly with others, decides on the purposes ("why") and means ("how") of the processing of personal data, the \aia focuses on the (traditionally single and) central provider of an AI system. 
However, since clients are autonomously in control of their data while the server is in control of the model, we see an inconsistency between what is controllable by the service provider and what he is responsible for. 
To close this gap, we need a two-pronged approach -- technical and legal ("how" \& "why"). 
Responsibility in FL should depend on the server's physical, technical, and legal ability to influence decentralized model training and configuration. 
This poses challenges. Unlike Article 26 and 28 GDPR, Article 8 et seq. of the \aia do not provide details on governance in networked processing environments like FL systems.

The data protection assessment of the FL lifecycle may be impacted if the FL server sets requirements for the clients, which may lead to the server being classified as the controller under the GDPR.
At first glance and from a strictly technical perspective, both the FL server and the FL clients fall under the provider concept of Article 3 GDPR, yet a “joint providership” (based on “joint controllership” under the GDPR, Article 26) does not exist under the \aia.
In fulfillment of its obligations under the \aia, in particular Article 10, the FL server can set far-reaching requirements for the FL clients concerning the training of models and handling of training data. These requirements could lead to the FL server being classified as a controller under data protection law within the means of Article 4 GDPR, while the FL client is classified as a mere processor within the scope of Article 28 GDPR. 

Thus, a key legal instrument for ensuring compliance with the \aia and GDPR  (Articles 26 \& 28) is likely the development of specific General Terms and Conditions binding for both FL server and clients. 
The server operator, as the service provider, has to oblige clients to provide sufficient reporting compliant with the \aia. This can be supported by cryptographic tools that minimize the need for trust among entities \cite{Nguyen2021}.

    \section{Considerations on Major Federated Learning Architectures}
    \label{sec:cross-dev_cross-sil_fl}
    \textit{Cross-silo FL may allow for more flexibility in system design and responsibility distribution between clients and server than cross-device FL.}

While \Cref{sec:technical_solutions} and \Cref{sec:regulatory_setup} are generally applicable to any FL application, there are two major system architectures that create further opportunities to organize responsibilities: cross-silo and cross-device training.



\subsection{Cross-Device Federated Learning}
Cross-device FL typically entails a large number of devices ($>1,000$). 
In such a setup, FL clients are characterized by having a very small number of local data samples and little participation time in the federated training process \cite{Hard2018}. 
This is a major challenge regarding client accountability and, ultimately, becomes problematic when a client should assume responsibility as a service provider.
Hence, for practical considerations, all responsibility has to be assumed by the server in the cross-device setting and there is practically no room for client-side responsibility and a strong need for tools and methods that allow the FL server to cover all compliance criteria. 
This implies that the runtime environment on clients must be as encapsulated as possible, coupled with strict terms of service agreements.

\subsection{Cross-Silo Federated Learning}
In contrast, in cross-silo settings, individual clients hold a significant amount of data and participate in multiple training rounds, and usually come with higher computational capabilities than in cross-device FL. 
Typically, cross-silo FL can involve large institutions such as hospitals \cite{Huang2022}, which themselves have a high commitment to regulatory compliance and take strong precautions regarding security and privacy protection. 
As such, it is an open research direction to explore the synergies between established institutional processes (e.g., medical record keeping) and the \aia requirements (e.g., on data transparency). 
The terms and conditions must be balanced between ensuring appropriate regulatory compliance and practical utility such that clients are incentivized to participate in training, and we can assume partial responsibility on the side of clients.
Such synergies could help better balance the service provider responsibilities and reduce costs for clients and the server, not only improving the economic viability of FL but also its ecological footprint.



    \section{Conclusion}
    \label{sec:conclusion}
    We study the FL life-cycle responsibilities under the \aia. 
We find client-side responsibility for numerous steps, which practically limits the applicability of FL to open up additional data silos that would benefit the training of foundation models. 
Yet, there are promising directions that deserve increased attention such that a server operator can become the \emph{service provider} without clients being required to assume extensive liability.
With this, one can drive the adoption of FL and help decrease data bias by directly relying on user data.
Further clarifying the outlined service provider question directly responds to the EU AI Office's call for contributions to help implement the \aia \cite{nature_call_ai_office_2024}. 


    \section*{Acknowledgements}
    This work is partially funded by the Bavarian Ministry of Economic Affairs, Regional Development and Energy (Grant: DIK0446/01).
    
    \bibliography{main}
    \bibliographystyle{icml2024}

\end{document}